\documentclass[]{bytedance_seed}

\usepackage[linesnumbered, ruled, vlined]{algorithm2e}
\usepackage{amsmath}
\usepackage{amssymb}
\usepackage{mathtools}
\usepackage{longtable}
\usepackage{booktabs}
\usepackage{array}
\usepackage{amsthm}
\usepackage{bbm}      
\usepackage{listings}
\usepackage{xcolor}   
\usepackage[most]{tcolorbox}
\usepackage{subcaption}
\usepackage[toc,page,header]{appendix}
\usepackage{minitoc}
\usepackage{prettyref}
\usepackage{tikz}
\usetikzlibrary{shapes, arrows.meta, positioning, calc, backgrounds, shadows}

\lstdefinestyle{mystyle}{
    backgroundcolor=\color{white},   
    commentstyle=\color{green!60!black},
    keywordstyle=\color{blue},
    numberstyle=\tiny\color{gray},
    stringstyle=\color{purple},
    basicstyle=\ttfamily\footnotesize,
    breakatwhitespace=false,         
    breaklines=true,                 
    captionpos=b,                    
    keepspaces=true,                 
    numbers=left,                    
    numbersep=5pt,                  
    showspaces=false,                
    showstringspaces=false,
    showtabs=false,                  
    tabsize=4
}

\newrefformat{eq}{(\ref{#1})}
\newrefformat{thm}{Theorem~\ref{#1}}
\newrefformat{th}{Theorem~\ref{#1}}
\newrefformat{chap}{Chapter~\ref{#1}}
\newrefformat{sec}{Section~\ref{#1}}
\newrefformat{seca}{Section~\ref{#1}}
\newrefformat{algo}{Algorithm~\ref{#1}}
\newrefformat{asmp}{Assumption~\ref{#1}}
\newrefformat{fig}{Fig.~\ref{#1}}
\newrefformat{tab}{Table~\ref{#1}}
\newrefformat{rmk}{Remark~\ref{#1}}
\newrefformat{clm}{Claim~\ref{#1}}
\newrefformat{def}{Definition~\ref{#1}}
\newrefformat{cor}{Corollary~\ref{#1}}
\newrefformat{lmm}{Lemma~\ref{#1}}
\newrefformat{prop}{Proposition~\ref{#1}}
\newrefformat{pr}{Proposition~\ref{#1}}
\newrefformat{property}{Property~\ref{#1}}
\newrefformat{app}{Appendix~\ref{#1}}
\newrefformat{apx}{Appendix~\ref{#1}}
\newrefformat{ex}{Example~\ref{#1}}
\newrefformat{exer}{Exercise~\ref{#1}}
\newrefformat{soln}{Solution~\ref{#1}}

\numberwithin{theorem}{section}

\newtheorem{example}{Example}

\title{LLM Swiss Round: Aggregating Multi-Benchmark Performance via Competitive Swiss-System Dynamics}
\author[1,\dagger]{\small Jiashuo Liu}
\author[1,2, \star]{\small Jiayun Wu}
\author[1]{\small Chunjie Wu}
\author[1]{\small Jingkai Liu}
\author[1]{\small Zaiyuan Wang}
\author[1]{\small Huan Zhou}
\author[1, \dagger]{\small  \\ Wenhao Huang}
\author[3]{\small Hongseok Namkoong}
\affiliation[1]{\small ByteDance Seed}
\affiliation[2]{\small Carnegie Mellon University}
\affiliation[3]{\small Columbia University}
\contribution[\dagger]{\small Corresponding author}
\contribution[\star]{\small Intern at ByteDance Seed}
\abstract{
The rapid proliferation of Large Language Models (LLMs) and diverse specialized benchmarks necessitates a shift from fragmented, task-specific metrics to a \textit{holistic, competitive ranking system} that effectively aggregates performance across multiple ability dimensions. 
Primarily using static scoring, current evaluation methods are fundamentally limited. They struggle to determine the proper mix ratio across diverse benchmarks, and critically, they fail to capture a model's dynamic competitive fitness or its vulnerability when confronted with sequential, high-stakes tasks.
To address this, we introduce the novel \textbf{Competitive Swiss-System Dynamics (CSD)} framework. CSD simulates a multi-round, sequential contest where models are dynamically paired across a curated sequence of benchmarks based on their accumulated win-loss record. And Monte Carlo Simulation ($N=100,000$ iterations) is used to approximate the \textit{statistically robust Expected Win Score} ($E[S_m]$), which eliminates the noise of random pairing and early-round luck. Furthermore, we implement a \textit{Failure Sensitivity Analysis} by parameterizing the per-round elimination quantity ($T_k$), which allows us to profile models based on their \textit{risk appetite}—distinguishing between robust generalists and aggressive specialists. We demonstrate that CSD provides a more nuanced and context-aware ranking than traditional aggregate scoring and static pairwise models, representing a vital step towards risk-informed, next-generation LLM evaluation.
}
\correspondence{Jiashuo Liu, Wenhao Huang at \email{\{liujiashuo.77, huang.wenhao\}@bytedance.com}}

\begin{document}
\maketitle
\begin{figure}
    \vspace{-0.8in}
    \centering
    \includegraphics[width=0.95\linewidth]{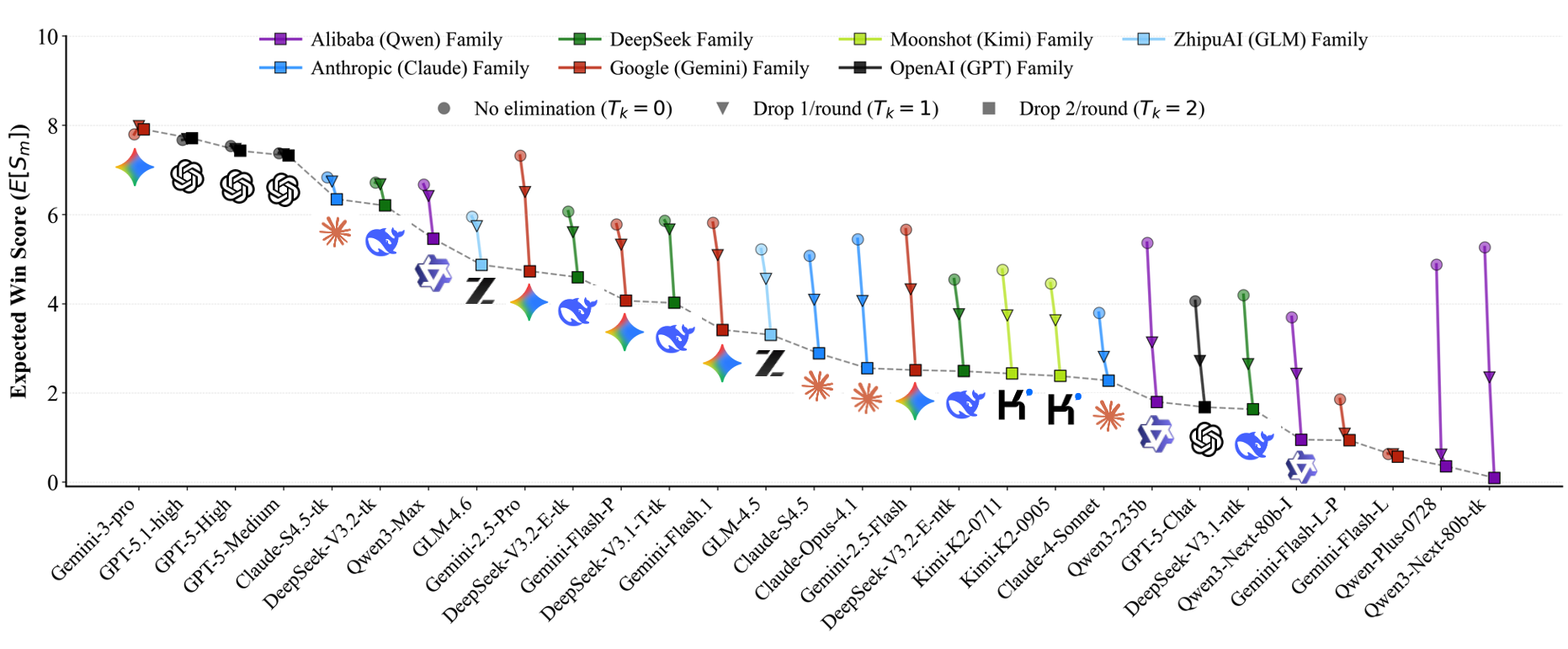}
    \vspace{-0.2in}
    \caption{Overall ranking of \textbf{29 advanced LLMs} across \textbf{38} recent widely-used and open-sourced benchmarks given by our Competitive Swiss-System Dynamics framework. Check if this aligns with your insights.}
    \vspace{-0.3in}
    \label{fig:placeholder}
\end{figure}

\newpage
\section{Introduction}
\label{sec:introduction}
The field of Artificial Intelligence has been rapidly transformed by the emergence and widespread deployment of Large Language Models (LLMs). These models exhibit remarkable capabilities across a multitude of tasks, spanning complex reasoning~\cite{rein2024gpqa, balunovicmathconstruct, bytedance_seed_2025_beyondaime, bytedance_seed_2025_scienceolympiad, fujisawaprocbench, phan2025humanity}, code generation~\cite{jainlivecodebench, quan2025codeelo, tambon2024assessing, zheng2025livecodebench, jimenezswe, zan2025multi}, and nuanced natural language understanding~\cite{wang2024mmlu, singh2024globalmmluunderstandingaddressing}.
Consequently, the development of robust evaluation methodologies has become paramount. However, the sheer diversity of benchmarks presents a challenge for practical model selection. In many downstream applications—ranging from selecting a backbone for autonomous agents to enterprise API procurement—practitioners face a singular decision: \emph{they must identify one model that is sufficiently robust to handle diverse, unpredictable workflows.} 
This challenge is exacerbated by the scale of modern LLM evaluation pipelines, which typically comprise hundreds of internal benchmarks. In such high-dimensional settings, manual inspection of individual metrics is practically infeasible.
Therefore, deriving a unified ranking from fragmented benchmarks is not merely a simplification, but a necessity for resource allocation and deployment decisions.
While standardized leaderboards attempt to provide this unified view (e.g., the ``Intelligence'' Index on \url{https://artificialanalysis.ai/}), they typically rely on simple aggregate scores. This approach often masks critical shortcomings, treating a model with high variance (excellent in one area, poor in another) as equivalent to a consistently reliable model. A truly deployable model must demonstrate uniformly high performance to be trusted in a multi-stage competitive environment.
To address the need for a reliable selection criterion, the key problem studied in this paper is:

\begin{tcolorbox}
\centering Given results on various benchmarks, how to synthesize a unified ranking that accurately reflects a model's general utility and robustness?
\end{tcolorbox}

\begin{figure}[htbp]
\centering
\begin{tikzpicture}[
    font=\sffamily\small,
    node distance=0.8cm, 
    box/.style={
        rectangle, rounded corners, minimum width=2.2cm, minimum height=1.0cm,
        draw=black!60, thick, drop shadow, align=center, fill=white
    },
    arrow/.style={
        ->, >=Stealth, thick, color=black!70
    },
    labeltext/.style={
        font=\bfseries\footnotesize, color=black!80
    }
]

\begin{scope}[local bounding box=leftpanel]
    \node[labeltext] at (0.5, 3.5) {A. Static Aggregation View};
    
    \draw[->, thick, black!50] (-1.5,0) -- (2.5,0); 
    \draw[->, thick, black!50] (-1.2,0) -- (-1.2,3) node[above] {\scriptsize Score};
    
    \node[anchor=south] at (-0.5, -0.5) {\scriptsize Instruction};
    \filldraw[fill=red!20, draw=red!80, thick] (-1,0) rectangle (0, 0.2); 
    \node[red!80, font=\bfseries\scriptsize] at (-0.5, 0.5) {Fail (0)};

    \node[anchor=south] at (1.5, -0.5) {\scriptsize Reasoning};
    \filldraw[fill=green!20, draw=green!80, thick] (1,0) rectangle (2, 2.8);
    \node[green!60!black, font=\bfseries\scriptsize] at (1.5, 3.0) {Excel (100)};

    \draw[dashed, blue, thick] (-1.2, 1.4) -- (2.5, 1.4);
    \node[blue, fill=white, inner sep=1pt, font=\bfseries\footnotesize, anchor=west] at (2.5, 1.4) {Avg: 50};
    
    \node[align=center, font=\footnotesize, text width=5.5cm, anchor=north] at (0.5, -0.8) {
        \textit{"Performs OK on average."}
    };
\end{scope}

\draw[thick, dotted, black!30] (3.8, -1.5) -- (3.8, 3.8);

\begin{scope}[shift={(8.5,0)}, local bounding box=rightpanel]
    \node[labeltext] at (0, 3.5) {B. Path-Dependent Reality};

    \node (step1) [box, fill=red!10, draw=red!80] at (0, 1.2) {
        \textbf{Step 1}\\Instruction
    };
    
    \node (input) [box, fill=gray!10, left=of step1] {User Input};
    
    \node (step2) [box, fill=gray!20, right=of step1, dashed, draw=gray!50] {
        \textbf{Step 2}\\Reasoning
    };
    
    \draw[arrow] (input) -- (step1);
    
    \draw[arrow, red, dashed] (step1) -- node[above, font=\bfseries\tiny,yshift=1pt] {BLOCKED} (step2);
    
    \node[red, font=\Huge, opacity=0.8] at (step1) {$\times$};
    
    \node (output) [below=0.8cm of step1, font=\bfseries\footnotesize, color=red] {Effective Utility: 0};
    \draw[->, red, thick, dotted] (step1.south) -- (output.north);

    \node[align=center, font=\footnotesize, text width=7cm, anchor=north] at (0, -0.8) {
        \textit{"Failure in Step 1 bottlenecks the entire pipeline."}
    };
\end{scope}

\end{tikzpicture}
\caption{The Illusion of Static Aggregation. (A) Averaging scores hides foundational failures. (B) In a realistic sequential workflow, failure in a foundational task (Step 1) blocks downstream capabilities (Step 2), illustrating the \textit{path dependency} of model performance.}
\label{fig:path_dependency_illustration}
\end{figure}
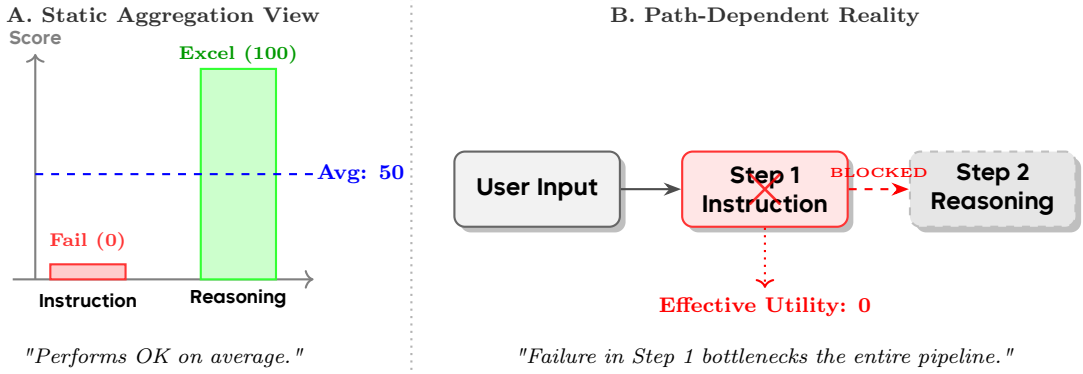

Current evaluation paradigms primarily rely on static aggregation, a method fundamentally challenged by the lack of an objective ground truth for task importance. We refer to this as the problem of \textit{arbitrary weighting}: when combining results from disparate benchmarks (e.g., math, coding, and safety), researchers must assign weights based on heuristics rather than data. Consequently, the final ranking becomes highly sensitive to these manual choices.
Furthermore, existing methods—whether based on simple averaging or static pairwise models like Elo~\cite{chiang2024chatbot}—fail to capture the \textit{path-dependent} nature of real-world model utility. In a static average, a failure in a foundational capability can be mathematically compensated by excellence in an advanced task. However, in practical deployment, capabilities are often sequential and interdependent.
Consider a typical agentic workflow illustrated in Figure~\ref{fig:path_dependency_illustration}: a model must first correctly parse a user's instruction (Step 1) before executing complex reasoning (Step 2). If a model fails the foundational Step 1, its potential proficiency in Step 2 is rendered irrelevant.
Static metrics obscure this distinction, treating the two capabilities as independent addends. We argue that a robust evaluation must bypass the arbitrary weighting problem and instead model evaluation as a dynamic competitive system—one that integrates cumulative pressure and structured elimination to reveal a model's true, risk-adjusted performance.

To bridge the gap between static aggregation and dynamic deployment, we introduce the \textbf{Competitive Swiss-System Dynamics (CSD)} framework for \textit{Holistic LLM Ranking}. Our framework simulates a multi-round environment where $M$ models are sequentially tested across $K$ distinct ability benchmarks using the Swiss-System pairing strategy. This approach introduces three primary innovations.
First, CSD resolves the arbitrary weighting problem by replacing heuristic coefficients with \textit{structural importance}. Unlike standard Elo implementations or static averages—which typically treat all benchmarks as independent data points—the Swiss-System enforces \textit{path dependency}. The impact (or ``weight'') of a benchmark is not assigned by the researcher, but emerges from the tournament dynamics: a failure in early rounds alters a model's pairing trajectory and maximum potential score. Thus, weighting becomes a function of competitive survival rather than subjective preference.
Second, to mitigate the variance inherent in tournament pairings, we leverage Monte Carlo Simulation ($N=100,000$ iterations) to derive a statistically robust Expected Win Score ($E[S_m]$). This metric represents a model's predicted cumulative victories, effectively eliminating the confounding influence of random pairing luck.
Finally, we integrate a structured elimination mechanism with a tunable parameter $T_k$ (the quantity of eliminated models in the lowest performance group). This facilitates a formal Failure Sensitivity Analysis, enabling us to profile models based on risk. Overall, our CSD framework prioritizes models that demonstrate consistent performance across benchmarks (i.e. \emph{Robust Generalist}), while heavily penalizing those with localized deficiencies (i.e. \emph{Aggressive Specialist}).

In summary, this paper makes the following contributions to LLM evaluation: 
\begin{itemize}
    \item A novel CSD framework that translates static multi-benchmark results into a dynamic, competition-based ranking, intrinsically avoiding the subjective weighting problem. 
    \item The $E[S_m]$ metric, a statistically robust and competition-aware score that provides a superior alternative to subjective aggregated scores.
    \item A quantifiable methodology for assessing models' Failure Sensitivity by analyzing $E[S_m]$’s behavior across varying elimination parameters $T_k$.
\end{itemize}
The remainder of this paper is organized as follows: 
\Cref{sec:methodology} details the formal structure of the CSD framework, including the Swiss-System rules, elimination mechanisms, and the Monte Carlo methodology. 
\Cref{sec:related_work} reviews related work in LLM benchmarking and competitive ranking systems. 
\Cref{sec:experiment} presents experimental results, comparing CSD rankings against traditional methods and demonstrating the Failure Sensitivity Analysis. 
\Cref{sec:conclusion} concludes the paper and discusses future research directions.
\section{Methodology}
\label{sec:methodology}

This section details the mathematical and conceptual foundations of the \textbf{Competitive Swiss-System Dynamics (CSD)} framework. 
We first establish the formal components of the competitive environment, justifying our design choices (\Cref{sec:components}). 
We then detail the stochastic state transition process (\Cref{sec:stochastic_process}), leading to our core argument for the computational intractability of an analytical solution (\Cref{sec:intractability}). 
This logically necessitates the Monte Carlo approximation (\Cref{sec:monte_carlo}) and enables our novel Failure Sensitivity Analysis (\Cref{sec:fsa}).

\subsection{Framework Components and Rationale}
\label{sec:components}

The CSD framework formally instantiates a multi-round, sequential competitive environment, engineered to dynamically evaluate LLMs under conditions of elevated stakes and cumulative performance pressure.

\textbf{The Necessity of Dynamic Competition in Deployment.}\quad Static evaluation, based on (heuristically) weighted averages, fails to adequately penalize critical performance gaps, which poses a significant risk in sequential industrial applications. 
The CSD framework is explicitly designed to model and mitigate this risk through structured competition.
We demonstrate the motivation through some real-world scenarios:
\begin{itemize}
    \item \textit{Supply Chain Automation:} Consider an LLM managing a supply chain, where the foundational task ($d_1$: parsing inventory manifests) precedes complex tasks ($d_k$: optimizing logistics routes). A model that fails $d_1$ but excels at $d_k$ is an unacceptable risk. CSD sequences $d_1$ first, forcing the fragile model into the minimum score group ($G_{\min}$), thereby exposing it to the \textit{probabilistic elimination risk} controlled by the parameter $T_k$.
    \item \textit{Financial Risk Assessment:} In loan underwriting, an LLM must possess strong capability in both numerical reasoning ($d_a$) and regulatory text interpretation ($d_b$). CSD uses its Swiss-System dynamic to pair high-scoring models on $d_a$ against strong performers on $d_b$. This ensures that final ranking reflects genuine \textit{full-spectrum competitive resilience}, rather than isolated proficiency in one area.
    \item \textit{Complex Code Generation:} A multi-step coding agent must correctly complete module $d_1$ to proceed to $d_2$. CSD’s \textit{path-dependent scoring} rewards models that successfully navigate the entire sequence. Consistent early wins accumulate score and provide a buffer, ensuring only models capable of sustaining performance survive the rigorous, later-stage competition.
\end{itemize}
Thus, CSD fundamentally shifts the evaluation focus from a theoretical average score to \textit{risk-adjusted fitness} for a sequential pipeline, ensuring that only models with verifiable, full-spectrum capability are rewarded with a high Expected Win Score ($\hat{E}[S_m]$).
We formalize the CSD framework in Algorithm~\ref{alg:csd_single_instance}.
Below, we will introduce our CSD framework in detail.

\subsubsection{The Pairwise Win-rate Tensor ($W$)}
The foundational data structure for our framework is the \textit{Pairwise Win-rate Tensor ($W$)}, an $M \times M \times K$ tensor where $M$ is the number of models and $K$ is the number of sequenced benchmarks.

\textbf{Conceptual Rationale.}\quad 
By abstracting raw performance scores (e.g., accuracy, perplexity) into binary win/loss outcomes, $W(i, j, k) \in \{0, 1\}$, the CSD framework inherently bypasses the critical and subjective problem of \textit{benchmark weighting}. As discussed in \Cref{sec:introduction}, deciding if a 10-point gain on a math benchmark is ``worth'' more than a 5-point gain on a coding benchmark is a subjective exercise that makes static-weighted averages fragile. 
In CSD, a win is a win. 
The ``importance'' of a benchmark is defined not by an \textit{a priori} subjective weight, but by its \textit{sequential position} in the contest and the competitive pressure at that stage. 
$W$ is pre-calculated from underlying model performance data.
Note that $W$ is only calculated once. This separation ensures that the iterative computational cost remains relatively low, as the expensive step is not repeated within the Monte Carlo sampling.

\subsubsection{The Swiss-System Pairing Engine}
\label{sec:swiss_system}
Given pre-defined $K$ sequenced benchmarks $\mathcal{D}_1, \dots, \mathcal{D}_K$, each game in our CSD framework contains multiple ($K$) rounds. 
In each round $k$, we first pair one model with another following the Swiss-System Pairing Engine, and then compare paired models according to their performances on the $k$-th round.

The core dynamic of the CSD is the Swiss-System pairing mechanism, which dictates \textit{who} competes with \textit{whom} in each round.

\textbf{Pairing Engine.}\quad In round $k$, the active models are first grouped based on their current cumulative win score $S_m(k-1)$. The system then pairs models within these score groups with the primary objective of matching opponents with identical scores. This mechanism ensures that model comparisons are concentrated among competitors of similar proven strength, thus maximizing the diagnostic value of each match.

The Swiss-System is chosen for its efficiency in ranking a large number of competitors with a limited number of rounds. Unlike a simple random-pairing tournament, its primary feature is \textit{dynamic strength-of-schedule matching}: models with similar cumulative scores are paired against each other. This ensures that:
\begin{enumerate}
    \item \textit{High-performing models} are rigorously tested against other high-performers, preventing them from achieving a high rank by only defeating weaker opponents.
    \item \textit{Low-performing models} are paired, allowing for a clearer differentiation at the bottom of the ranking, which is crucial for our elimination mechanism.
\end{enumerate}

\textbf{Zero-Point Bye Rule.}\quad 
A critical, non-standard design choice in our CSD framework is the \textit{Zero-Point Bye} rule. In traditional chess tournaments, a bye (receiving no opponent) may grant 1 or 0.5 points. We explicitly assign 0 points. 
This is because our objective is to measure \textit{competitive success}. 
A bye is a product of random chance (an odd number in a score group $G_s(k)$), not a competitive victory. 
This design choice ensures that $E[S_m]$ exclusively reflects accumulated \textit{competitive} victories, not passive, luck-based score inflation.
Furthermore, as detailed in Section~\ref{sec:monte_carlo}, the implementation of Monte Carlo Simulation---which involves repeating the contest $N$ times---effectively mitigates the influence of single-instance stochasticity (e.g., random pairing or elimination) to yield a statistically robust Expected Win Score ($\hat{E}[S_m]$).

\subsubsection{The Structured Elimination Mechanism}
The \textit{Structured Elimination Mechanism} is the CSD's core component for modeling risk and penalizing performance fragility. 
After each round $k$, a set of models ($T_k$ models) are permanently removed from the contest.

\textbf{Conceptual Rationale.}\quad 
This mechanism models the real-world deployment-cycle reality that models exhibiting significant failures are often ``eliminated'' from consideration. 
By targeting the \textit{minimum score group ($G_{\min}$)}, we ensure that elimination pressure is applied only to the models demonstrating the poorest relative performance in the contest up to that point. 
The parameter \textit{$T_k$ (Elimination Count)} thus acts as a tunable ``penalty for failure'' or ``pressure'' of the contest.

\subsection{Formal Stochastic Process}
\label{sec:stochastic_process}

We now formally define the state transition from round $k$ to $k+1$.
Let $S_m(k)$ denotes the score of model $m$ after $k$ rounds.
\begin{itemize}
    \item \textbf{State $\mathcal{X}_k = (\mathcal{M}_k, \vec{S}(k-1))$}: The set of active models $\mathcal{M}_k$ and their corresponding cumulative score vector $\vec{S}(k-1)$.
    
    \item \textbf{Phase 1: Grouping}: Models in $\mathcal{M}_k$ are partitioned into disjoint score groups $G_s(k) = \{m \in \mathcal{M}_k \mid S_m(k-1) = s\}$. We let $n_s(k) = |G_s(k)|$.
    
    \item \textbf{Phase 2: Scoring ($I_m(k)$)}: A model $m$'s score for the round, $I_m(k) \in \{0, 1\}$, is determined. Its conditional expectation, $E[I_m(k) \mid \mathcal{X}_k]$, given it is in group $G_s(k)$, is:
    \begin{equation}    
    E[I_m(k) \mid \mathcal{X}_k] = 
    \begin{cases} 
    \frac{1}{n_s(k)-1} \sum_{j \in G_s(k) \setminus \{m\}} W(m, j, k) & \text{if } n_s(k) \text{ is even}; \\ 
    \left(1 - \frac{1}{n_s(k)}\right) \cdot \left[ \frac{1}{n_s(k)-1} \sum_{j \in G_s(k) \setminus \{m\}} W(m, j, k) \right] & \text{if } n_s(k) \text{ is odd}. 
    \end{cases}
    \end{equation}
    
    \textbf{Interpretation}: This equation quantifies the expected gain for $m$ in round $k$. If $n_s(k)$ is odd, $m$ has a $1/n_s(k)$ chance of receiving a bye (score 0) and a $(1 - 1/n_s(k))$ chance of playing. If it plays, it faces one of the $n_s(k)-1$ others, and its expected score is its average win-rate against them on benchmark $d_k$.
    
    \item \textbf{Phase 3: Elimination}: Models update their scores $S_m(k) = S_m(k-1) + I_m(k)$. The new minimum group $G_{\min}(k)$ (size $n_{\min}(k)$) is identified. The probability of elimination for a model $m$ is:
    \begin{equation}    
    P_{\text{Elim}}(m, k) = \mathbf{1}(m \in G_{\min}(k)) \cdot \frac{T_k}{\max(T_k, n_{\min}(k))}.
    \end{equation}
    The surviving set $\mathcal{M}_{k+1}$ is formed, completing the transition.
\end{itemize}

\begin{small}
\begin{algorithm}[H]
\DontPrintSemicolon
\SetKwInOut{Input}{Input}
\SetKwInOut{Output}{Output}
\SetKwFunction{PairSwiss}{SwissPairing}
\SetKwFunction{Eliminate}{ApplyElimination}

\Input{
    Set of Models $\mathcal{M} = \{m_1, \dots, m_M\}$; 
    Sequenced Benchmarks $\mathcal{D} = \{d_1, \dots, d_K\}$; \\
    Win-rate Tensor $W(i, j, k)$ (where $k$ corresponds to $d_k$); \\
    Number of Rounds $K$; 
    Elimination Parameter $T_k$ (for each round $k$);
}
\Output{
    Final Score Vector $\vec{S}(K)$; 
    Final Model Ranking;
}

\BlankLine
$\mathcal{M}_{\text{active}} \leftarrow \mathcal{M}$ \tcp{Set of active models}
$\vec{S}(0) \leftarrow \vec{0}$ \tcp{Initialize all cumulative scores to zero}

\For{$k = 1$ \KwTo $K$}{ 
    \textbf{Current Benchmark $d_k$ is used for round $k$ contests;}
    
    \emph{// Phase 1: Grouping and Pairing} \;
    Partition $\mathcal{M}_{\text{active}}$ into score groups $G_s(k)$ based on $\vec{S}(k-1)$ \;
    $Pairs_k \leftarrow \emptyset$ \;
    
    \For{each score group $G_s(k)$}{ 
        $Pairs_{k, s} \leftarrow \PairSwiss(G_s(k))$ \tcp{random pairs}
        $Pairs_k \leftarrow Pairs_k \cup Pairs_{k, s}$ \;
    }

    \BlankLine
    \emph{// Phase 2: Contest Execution and Scoring on $d_k$} \;
    \For{$(m_i, m_j)$ in $Pairs_k$}{ 
        $I_{m_i}(k) \leftarrow W(i, j, k)$ \tcp{Look up win/loss for $d_k$}
        $I_{m_j}(k) \leftarrow 1 - I_{m_i}(k)$ \;
    }
    \For{each model $m$ that received a Bye}{
        $I_m(k) \leftarrow 0$ \tcp{\textbf{Zero-Point Bye Rule}}
    }

    \BlankLine
    \emph{// Phase 3: Score Update and Elimination} \;
    \For{each $m \in \mathcal{M}_{\text{active}}$}{
        $S_m(k) \leftarrow S_m(k-1) + I_m(k)$ \tcp{Update cumulative score}
    }
    
    $G_{\min}(k) \leftarrow \{m \in \mathcal{M}_{\text{active}} \mid S_m(k) = \min(\vec{S}(k))\}$ \tcp{Identify minimum score group}
    
    $\mathcal{M}_{\text{active}} \leftarrow \Eliminate(\mathcal{M}_{\text{active}}, G_{\min}(k), T_k)$ \tcp{Randomly remove $T_k$ models from $G_{\min}(k)$}
    
    \If{$|\mathcal{M}_{\text{active}}| < 2$}{
        \textbf{break} \tcp{Competition ends if fewer than 2 models remain}
    }
}
\caption{Single Instance of Competitive Swiss-System Dynamics (SingleInstanceCSD)}
\label{alg:csd_single_instance}
\end{algorithm}
\end{small}

\SetKwFunction{CSD}{SingleInstanceCSD} 
\begin{small}
\begin{algorithm}[H]
\DontPrintSemicolon
\SetKwInOut{Input}{Input}
\SetKwInOut{Output}{Output}

\Input{
    Set of Models $\mathcal{M}$; 
    Sequenced Benchmarks $\mathcal{D}$;  
    Win-rate Tensor $W$; 
    Number of Rounds $K$; 
    Elimination Parameter $T_k$; 
    \textbf{Number of Monte Carlo Iterations $N$};
}
\Output{
    Estimated Expected Win Score Vector $\hat{\vec{E}}[\vec{S}]$; 
    Statistically Robust Model Ranking;
}

\BlankLine
$M \leftarrow |\mathcal{M}|$ \tcp{Number of models}
$\vec{S}_{\text{total}} \leftarrow \vec{0}_{M}$ \tcp{Initialize total score accumulator for all models}

\For{$i = 1$ \KwTo $N$}{
    \emph{// Run a single, stochastic CSD competition instance} \;
    \textbf{$\vec{S}^{(i)}(K) \leftarrow \CSD(\mathcal{M}, \mathcal{D}, W, K, T_k)$ \tcp{Returns final score vector}}
    \emph{// Accumulate the final scores} \;
    \For{each model $m_j \in \mathcal{M}$}{
        $S_{\text{total}, j} \leftarrow S_{\text{total}, j} + S_{m_j}^{(i)}(K)$ \;
    }
}

\BlankLine
\emph{// Estimate the Expected Win Score $\hat{E}[S_m]$} \;
\For{each model $m_j \in \mathcal{M}$}{
    $\hat{E}[S_{m_j}] \leftarrow S_{\text{total}, j} / N$ \tcp{Sample mean over all $N$ trials}
}
$\hat{\vec{E}}[\vec{S}] \leftarrow \{\hat{E}[S_{m_j}]\}_{j=1}^{M}$ \;
\caption{Monte Carlo Approximation of CSD Expected Win Score ($\hat{E}[S]$)}
\label{alg:monte_carlo}
\end{algorithm}
\end{small}

\subsection{Intractability of the Analytical Solution}
\label{sec:intractability}

Our objective is to compute the \textbf{Expected Win Score ($E[S_m]$)} for each model $m$ over the $K$ rounds. By the linearity of expectation,
\begin{equation}    
E[S_m(K)] = \sum_{k=1}^{K} E[I_m(k)].
\end{equation}

To compute $E[I_m(k)]$, one must use the law of total expectation:
\begin{equation}    
E[I_m(k)] = \sum_{\text{all possible } \mathcal{X}_k} E[I_m(k) \mid \mathcal{X}_k] \cdot P(\mathcal{X}_k).
\end{equation}
This analytical solution is \textit{computationally intractable}. 
The intractability arises from the \textit{path-dependent nature} of the process, leading to a \textit{combinatorial explosion} of the state space. 
The state in round $k$, $\mathcal{X}_k$, depends on the entire history of stochastic events:
\begin{enumerate}
    \item \textit{Pairing Stochasticity}: The random pairings within $G_s(j)$ for all $j < k$.
    \item \textit{Elimination Stochasticity}: The random eliminations from $G_{\min}(j)$ for all $j < k$.
\end{enumerate}
The number of possible ``contest histories'' grows exponentially, making the direct computation of the probability $P(\mathcal{X}_k)$ for every possible state $\mathcal{X}_k$ infeasible for any non-trivial $M$, $K$, and $W$.

\subsection{Monte Carlo Approximation of $E[S_m(K)]$}
\label{sec:monte_carlo}

Given the intractability of the analytical solution, we must approximate $E[S_m(K)]$ using \textit{Monte Carlo Simulation}.
\begin{enumerate}
    \item \textit{Simulation}: We simulate the entire $K$-round CSD contest $N$ times (e.g., $N=10,000$), where each simulation $i$ is a full ``path realization'' from $\mathcal{X}_1$ to $\mathcal{X}_{K+1}$.
    \item \textit{Estimation}: The estimator $\hat{E}[S_m(K)]$ is the sample mean of the final scores $S_{m}^{(i)}(K)$:
    \begin{equation}    
    \hat{E}[S_m(K)] = \frac{1}{N} \sum_{i=1}^{N} S_{m}^{(i)}(K).
    \end{equation}
\end{enumerate}
\textbf{Conceptual Interpretation of $E[S_m]$.}\quad 
By the law of large numbers, $\hat{E}[S_m] \to E[S_m]$. This metric is far richer than a simple average score. It represents a model's \textit{expected cumulative victories \textit{given} its ability to survive the sequential elimination pressures} of the CSD. It is a holistic metric that intrinsically blends a model's raw win-rate (from $W$) with its robustness against failure (its ability to stay out of $G_{\min}$) and its resilience to random chance (luck in pairing and elimination draws).
The whole procedure is shown in Algorithm~\ref{alg:monte_carlo}.

\subsection{Failure Sensitivity Analysis (FSA)}
\label{sec:fsa}

The CSD framework's true diagnostic power is unlocked by the \textbf{Failure Sensitivity Analysis (FSA)}. This analysis elevates our framework from a simple ranking tool to a \textit{diagnostic profiling system}.

Specifically, a single $\hat{E}[S_m]$ score (at a fixed $T_k$) provides a ranking, but the \textit{function} $\hat{E}[S_m](T_k)$ provides a \textit{model risk profile}. 
This profile reveals the trade-offs between a model's aggressive, high-scoring potential and its defensive robustness, providing a multi-dimensional basis for model selection that aligns with specific deployment risk-tolerances (e.g., ``Is it better to have a model that is excellent at 9 tasks but fails 1, or one that is good at all 10?'').

\textbf{Procedure.}\quad 
For simplicity, we fix the elimination count $T_k$ for each round  (e.g., $T_k \in \{1, 2, 3\}$). 
We then run the full $N$-iteration Monte Carlo simulation for a range of $T_k$ values to trace the $E[S_m]$ curve.
We define a model's \textbf{Sensitivity Coefficient ($\Lambda_m$)} as the empirical derivative (e.g., slope of a linear regression) of this function:
\begin{equation}
\label{equ:sensitivity}
    \Lambda_m \approx \frac{\Delta \hat{E}[S_m]}{\Delta T_k}.    
\end{equation}
This coefficient $\Lambda_m$ allows us to classify models (see~\Cref{fig:two-dimension}):
\begin{itemize}
    \item \textbf{Aggressive Specialist ($\Lambda_m \ll 0$).}\quad A model with a highly negative slope. Its high $E[S_m]$ at low $P_T$ (low penalty) reveals its ``specialist'' nature, but this score collapses as the penalty $P_T$ increases, exposing its fragility to ``short boards''.
    \item \textbf{Robust Generalist ($\Lambda_m \approx 0$).}\quad A model with a near-zero slope. Its $E[S_m]$ is stable and insensitive to elimination pressure, indicating it rarely, if ever, falls into the $G_{\min}$ group.
\end{itemize}
\section{Related Work}
\label{sec:related_work}

\textbf{Existing Evaluation Paradigms: Pointwise Evaluation.}\quad LLM evaluation has historically been dominated by \textit{pointwise benchmarking} (e.g., HELM, GLUE, MMLU). This paradigm assesses model performance in isolation, yielding a scalar score (e.g., accuracy, F1-score) per task. While essential for measuring task-specific proficiency, pointwise methods inherently fail to capture the \textit{relative competitive strength} between models, making holistic comparisons challenging. Furthermore, any aggregated ranking derived from pointwise scores relies on arbitrary, subjective weighting across diverse benchmarks, a fundamental flaw addressed in our Introduction. This reliance on static, independent metrics led to the development of methods that focus on relative comparison.

\textbf{Pairwise Ranking: Elo and Bradley-Terry Models}.\quad To overcome the limitations of pointwise scoring, the research community adopted \textit{pairwise ranking models}, most notably those based on the \textit{Bradley-Terry (BT) model} and the \textit{Elo rating system}. Systems like the Chatbot Arena (LLM Arena) utilize the Elo methodology---derived from human preference data via crowd-sourcing---to generate a general, single-valued skill rating for each model. The mathematical rigor of Elo and BT lies in their probabilistic foundation, which translates win-loss records into a latent ``skill parameter'' that best explains the observed outcomes. These models are crucial for providing a single, universally comparable skill score that is independent of the dataset used, a significant advantage over simple averaging.

\textbf{CSD Framework vs. Static Pairwise Ranking (Elo)}.\quad 
While acknowledging the statistical rigor of Elo/BT systems, our Competitive Swiss-System Dynamics (CSD) framework diverges fundamentally in its objective and structure. The key differences can be summarized as a shift from \textit{static, general-purpose ranking} to \textit{dynamic, contest-specific profiling}:

\begin{enumerate}
    \item \textbf{Objective: General Skill vs. Competitive Fitness}: Elo aims to compute a model's \textit{universal, equilibrium skill rating ($R$)}---a score that predicts future win probability independent of the competition format. In contrast, CSD computes the \textit{Expected Win Score ($E[S_m]$)} within a \textit{prescribed, high-stakes competition structure}. $E[S_m]$ measures a model's fitness \textit{for that specific contest}, intrinsically folding the penalty of sequential failure and the reward of surviving elimination into the final metric.
    
    \item \textbf{Information Scope and Dynamics}: Elo and BT models are typically designed to process large volumes of independent pairwise outcomes (human votes or standardized head-to-head results). They do not account for \textit{path dependency}. CSD, however, leverages the \textit{Swiss-System dynamic}, where pairing in round $k$ is determined by the cumulative history of wins $S_m(k-1)$. This dynamic ensures high performers face progressively harder opponents, providing a far more realistic simulation of sustained competition.
    
    \item \textbf{Risk Quantification (Failure Sensitivity)}: Traditional pairwise models yield only one dimension: skill ($R$). The CSD framework introduces the \textit{Failure Sensitivity Analysis (FSA)} via the elimination parameter $T_k$. This allows CSD to quantify a model's \textit{risk profile}---its vulnerability to being eliminated due to a single ``short board''---a diagnostic capability entirely absent in standard Elo systems. Our method thus offers a crucial \textit{second dimension of evaluation} for deployment safety and reliability.
\end{enumerate}

\textbf{Contest Simulation and Next-Generation Evaluation.}\quad 
Our work contributes to the emerging field of \textit{Contest-Based Evaluation}. Existing work on applying tournament structures to AI (e.g., in game theory and multi-agent systems) often focuses on optimizing pairing strategies for efficiency. CSD, by contrast, focuses on using the \textit{tournament structure itself as a diagnostic tool}. By combining the statistical robustness of \textit{Pairwise Data ($W$ matrix)} with the \textit{dynamic structure of the Swiss-System} and the rigorous sampling of \textit{Monte Carlo simulation}, CSD represents a vital step toward a \textit{next-generation evaluation framework}. This framework moves beyond passive measurement, offering an \textit{actionable, risk-informed, and context-aware} methodology for ranking LLMs suitable for specific deployment pipelines.
\section{Experiment} \label{sec:experiment} 

This section presents the experimental evaluation of our CSD system through two key analyses: (1) an overall comparative analysis of how the most advanced LLMs ranking across multiple established benchmarks, and (2) an in-depth examination of LLM ranking within a specific, individual benchmark.

Throughout this section, we examine the most advanced LLMs (\textbf{29} LLMs in total), including:
\begin{itemize}
    \item \textbf{Google (Gemini Series):} Gemini-3-pro, Gemini-2.5-Pro, Gemini-2.5-Flash, Gemini-2.5-Flash.1, Gemini-2.5-Flash-Lite, Gemini-2.5-Flash-Lite-Preview-2509, Gemini-2.5-Flash-Preview-2509.
    \item \textbf{OpenAI (GPT-5 Series):} GPT-5.1-high, GPT-5-chat, GPT-5-high, GPT-5-medium.
    \item \textbf{Anthropic (Claude Series):} Claude-Opus-4.1-nothinking, Claude-4-Sonnet-nothinking, Claude-Sonnet-4.5-nothinking, Claude-Sonnet-4.5-thinking.
    \item \textbf{Alibaba (Qwen Series):} Qwen-plus-0728, Qwen3-next-80b-a3b-thinking, Qwen3-next-80b-a3b-instruct, Qwen3-235b-a22b-instruct-2507, Qwen3-max-0923.
    \item \textbf{DeepSeek:} DeepSeek-V3.1-Terminus-nothinking, DeepSeek-V3.1-Terminus-thinking, DeepSeek-V3.2-Exp-nothinking, DeepSeek-V3.2-Exp-thinking, DeepSeek-V3.2-thinking.
    \item \textbf{Zhipu AI (Z.ai / GLM Series):} GLM-4.5, GLM-4.6.
    \item \textbf{Moonshot AI (Kimi Series):} Kimi-K2-0711, Kimi-K2-0905.
\end{itemize}

\subsection{Overall Ranking of Advanced LLMs Across 38 Benchmarks} 
\label{subsec:overall-ranking}
As previously discussed, the ultimate goal of the CSD framework is to derive a holistic LLM ranking across various benchmarks. To achieve this, we perform a multi-round, sequential contest among the models.

\textbf{Benchmarks.}\quad 
We evaluate model performance using a comprehensive suite of 38 recent, widely-used, and open-source benchmarks. 
Our selection methodology aims to reflect a contemporary and representative ranking aligned with community needs, focusing on two criteria: (1) \emph{widespread adoption} and (2) \emph{recent development}.
These benchmarks span 6 high-level capability categories: basic knowledge, reasoning, instruction following, coding, agent capabilities, and factuality.
For a more granular and sequential analysis (corresponding to our 12-round contest structure), these are further organized into 12 sequential sub-categories. Sub-categories are delineated based on either the difficulty spectrum or the scope of the respective benchmarks, ensuring a structured progression of challenges.
Detailed information regarding the sequential arrangement and mapping is provided in \Cref{tab:sequencial benchmarks}.

\begin{table}[h]
\centering
\caption{Sequencial benchmarks used in our CSD framework.}
\label{tab:sequencial benchmarks}
\begin{tabular}{>{\centering\arraybackslash}p{2.0cm} c c p{7cm}} 
\toprule
\textbf{Category} & \textbf{Round} & \textbf{Sub-Category} & \textbf{Cleaned Dataset Names} \\
\midrule
\multirow{2}{=}{\centering Basic Knowledge} & 1 & Foundational Knowledge & MMLU~\cite{hendrycksmeasuring}, MMLU-pro~\cite{wang2024mmlu}, SuperGPQA~\cite{team2025supergpqa}, SimpleQA~\cite{wei2024measuring}, ChineseSimpleQA~\cite{he2024chinese} \\
\cmidrule{2-4}
& 2 & Others & ArenaHard(v2)~\cite{li2024live}, GMMLU(lite)~\cite{singh2024globalmmluunderstandingaddressing}, MMMLU~\cite{hendrycksmeasuring} \\
\midrule
\multirow{2}{=}{\centering Instruction Following} & 3 & Basic Instruction Following & IFEval~\cite{zhou2023instruction}, MulDimIF~\cite{ye2025multi} \\
\cmidrule{2-4}
& 4 & Complex Instruction Following & EIFBench~\cite{zou2025eifbench}, MultiChallenge~\cite{sirdeshmukh2025multichallenge}, MARS-Bench~\cite{yang2025mars} \\
\midrule
\multirow{4}{*}{\centering Reasoning} & 5 & Logic Puzzles & ARC-AGI-2~\cite{chollet2025arc}, ProcBench~\cite{fujisawaprocbench}, KORBENCH~\cite{makor} \\
\cmidrule{2-4}
& 6 & Mathematics & AIME24, AIME25\footnote{\url{https://artofproblemsolving.com/wiki/index.php/AIME_Problems_and_Solutions}}, BeyondAIME~\cite{bytedance_seed_2025_beyondaime}, MathConstruct~\cite{balunovicmathconstruct} \\
\cmidrule{2-4}
& 7 & Other Disciplines & GPQA(diamond)~\cite{rein2024gpqa}, ScienceOlympiad~\cite{bytedance_seed_2025_scienceolympiad}, Phybench~\cite{qiu2025phybench}, PHYSICS~\cite{feng2025physics}\\
\cmidrule{2-4}
& 8 & Algorithms & LiveCodeBench~\cite{jainlivecodebench}, CodeForces~\cite{quan2025codeelo}, HardEval~\cite{tambon2024assessing}\\
\midrule
\multirow{1}{*}{\centering Coding} & 9 & Software Engineering & SWE-Bench~\cite{jimenezswe}, Multi-SWE-Bench~\cite{zan2025multi}, Terminal Bench\footnote{\url{https://www.tbench.ai/}} \\
\midrule
\multirow{2}{*}{\centering Agent} & 10 & Search Agent & GAIA~\cite{mialon2023gaia}, HLE~\cite{phan2025humanity}, BrowseComp~\cite{wei2025browsecomp}, BrowseComp-zh~\cite{zhou2025browsecomp} \\
\cmidrule{2-4}
& 11 & Tool Use & $\tau^2$-bench~\cite{barres2025tau} \\
\midrule
\multirow{1}{=}{\centering Factuality} & 12 & Factual Hallucinations & FActScore~\cite{min2023factscore}, LongFact (object)~\cite{NEURIPS2024_937ae0e8}, LongFact (concept)~\cite{NEURIPS2024_937ae0e8} \\
\bottomrule
\end{tabular}
\end{table}

\textbf{Order of Benchmarks.}\quad 
As shown in~\Cref{tab:sequencial benchmarks}, the sequencing of these benchmarks follows a robust two-part rationale: moving from general to specific coverage, and progressing from fundamental to challenging tasks.
For instance, core abilities such as general knowledge and instruction following form the essential prerequisites upon which complex capabilities—like reasoning, coding, and autonomous agent skills—are built. We therefore place these foundational capacities at the beginning of our Swiss System schedule.

This specific sequence is critical within our CSD framework. Recall that CSD involves elimination after each round, and the early ranking is determined by the initial, fundamental tasks. While a model might excel at an advanced task (e.g., coding or agent work), poor performance on the fundamental steps (general knowledge and instruction following) indicates a severe deficiency. A model that cannot correctly understand our demands or lacks basic factual grounding cannot be expected to succeed reliably in complex tasks, regardless of its specialized potential.
By prioritizing the foundation, the CSD framework \emph{naturally} imposes a higher implied weight on models that demonstrate strong performance in basic competencies early on. This ensures that models which pass the initial low-level filter are inherently more reliable and capable of handling the demands of subsequent, specialized tasks.

\begin{figure}
    \centering
    \includegraphics[width=\linewidth]{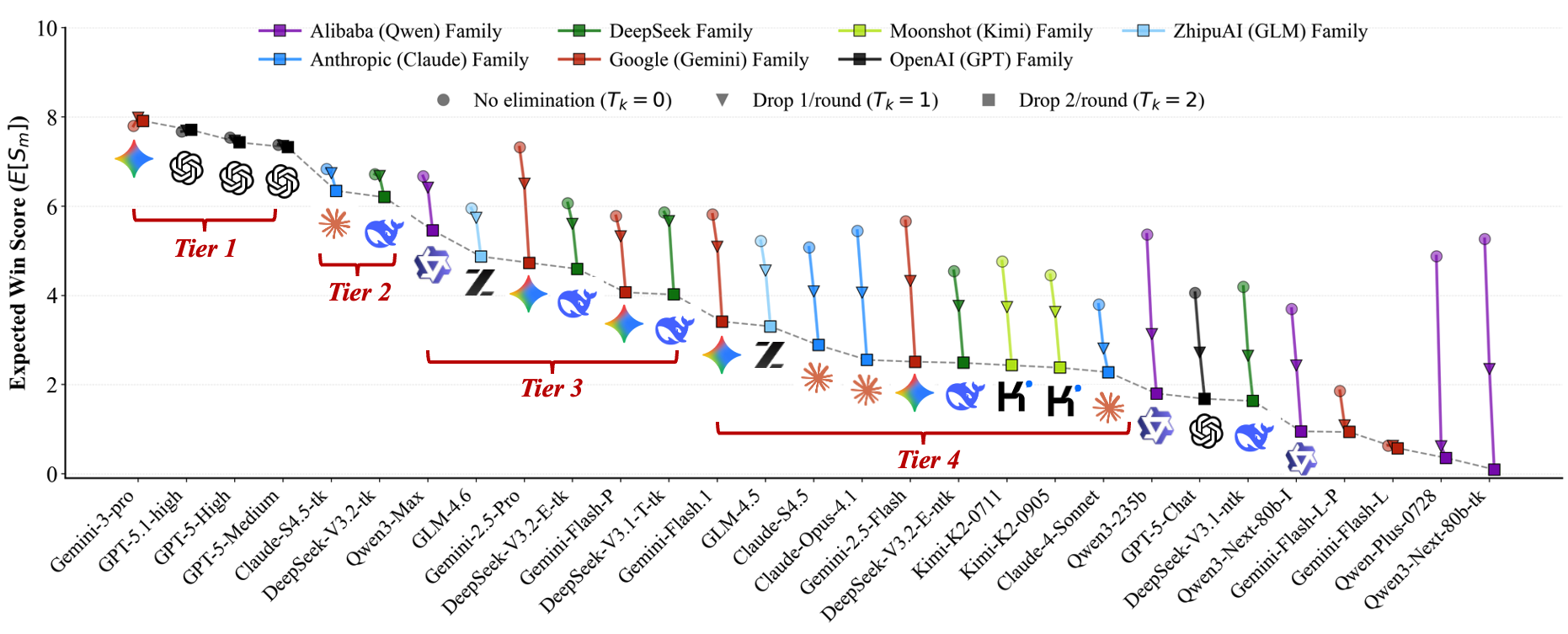}
    \caption{Overall ranking of 29 advanced LLMs across 38 recent widely-used and open-sourced benchmarks given by our CSD framework, highlighting the models organized into four tiers.}
    \label{fig:tier}
\end{figure}

\textbf{Finding 1: The Overall Ranking Aligns with Consensus and Reveals Four Performance Tiers.}\quad 
The overall ranking of the 29 advanced LLMs is visually represented in \Cref{fig:tier}, establishing the general capability hierarchy among these state-of-the-art models.
\Cref{fig:tier} reveals three distinct layers of top-performing models. 
\begin{itemize}
    \item The \emph{first tier} is led by Gemini-3-pro, GPT-5.1-High, GPT-5-High, and GPT-5-Medium which demonstrate highly similar performance. Critically, all these four models exhibit minimal degradation in score as the elimination pressure (models dropped per round) increases, indicating exceptional robustness and general competence. These three LLMs collectively form the first tier, representing the most general and resilient models currently available.
    \item The \emph{second tier} of models is clearly demarcated, including Claude-Sonnet-4.5-thinking and DeepSeek-V3.2-thinking. Similar to the first tier, these two models exhibit relatively robust performance and general competence.
    \item The \emph{third tier} of models is led by Qwen3-Max. It is closely followed by a highly competitive group: GLM-4.6, Gemini-2.5-pro, DeepSeek-V3.2-Exp-Thinking, Gemini-2.5-Flash-Preview, and DeepSeek-V3.1-Terminus-Thinking. This cluster highlights a significant trend: the rapid performance improvement of Chinese models (GLM, DeepSeek, Qwen). Their scores demonstrate that they are effectively closing the performance gap with Gemini-2.5-Pro (and some have surpassed it), which previously defined the performance frontier (SOTA) in many benchmarks and remains a highly robust contender. This tier, therefore, illustrates the accelerating global competition and the quickly evolving landscape of general LLM capabilities.
    \item The \emph{fourth tier} contains models with suboptimal operational or architectural trade-offs, thus establishing a clear performance degradation. 
    This cohort primarily consists of previous model iterations (e.g., GLM-4.5), lightweight or efficiency-focused versions (e.g., Gemini-2.5-Flash-Lite, Gemini-2.5-Flash), and base models lacking external ``thinking'' or planning mechanisms (e.g., Kimi-K2-0905, DeepSeek-V3.2-Exp-NonThinking). This concentration confirms the strong relationship between model scale/architectural complexity and overall performance robustness under competitive evaluation.
\end{itemize}

\begin{figure}
    \centering
    \includegraphics[width=\linewidth]{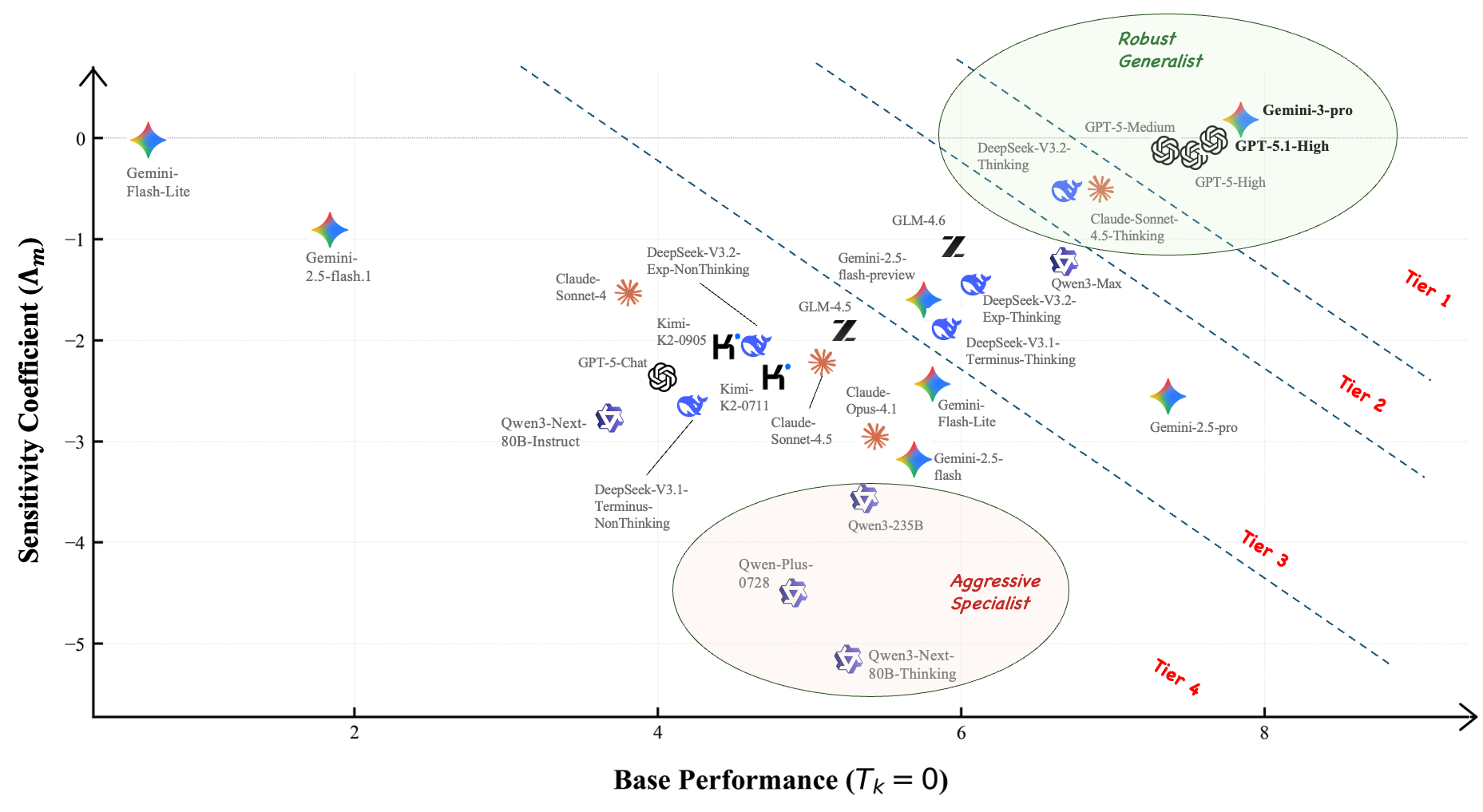}
    \caption{Base Performance and Sensitivity Coefficient of 29 Models. The base performance is the average model score given by our CSD framework when there is no model elimination ($T_k=0$). The sensitivity coefficient is the gradient in average score when $T_k$ increases from 0 to 2, calculated as $(E[S_m|T_k=2] - E[S_m|T_k=0])/2$. A more negative sensitivity coefficient indicates greater model shortcomings or susceptibility to elimination.}
    \label{fig:two-dimension}
\end{figure}

\textbf{Finding 2: Robust Generalists vs. Aggressive Specialists.}
To better understand how model performance is affected by competitive pressure, we define the performance drop as:
\begin{equation}
    \Delta E[S_m]\coloneqq E[S_m|T_k=2]-E[S_m|T_k=0] = 2\cdot \Lambda_m,
\end{equation}
which calculates the decrease in the average score when the system shifts from a baseline state ($T_k=0$, no elimination) to the $T_k=2$ condition (dropping two models per round), and $\Lambda_m$ is the sensitivity coefficient defined in~\Cref{equ:sensitivity}.
In~\Cref{fig:two-dimension}, we plot the base performance $E[S_m|T_k=0]$ against the sensitivity coefficient, where models positioned further to the right exhibit better overall performance, while models positioned higher up demonstrate fewer shortcomings relative to the group.
From~\Cref{fig:two-dimension}, we find that the three tiers observed in~\Cref{fig:tier} are clear.

As introduced in~\Cref{sec:fsa}, we conduct failure sensitivity analysis, where we define two distinct model behaviors based on their performance and resilience: the Robust Generalist and the Aggressive Specialist.
A Robust Generalist is characterized by high overall performance coupled with minimal shortcomings. Conversely, an Aggressive Specialist also achieves high overall performance but exhibits significant shortcomings (i.e., low robustness).
As visualized in Figure~\ref{fig:two-dimension}, models such as Gemini-3-pro, GPT-5.1-High, GPT-5-High, GPT-5-Medium, Claude-Sonnet-4.5-Thinking, and DeepSeek-V3.2-thinking exemplify the robust generalist behavior. Conversely, Qwen-3-235B, Qwen-Plus-0728, and Qwen-Next-80B-Thinking are grouped as aggressive specialists.

\subsection{Sensitivity Analysis of CSD Framework}
Following the overall ranking analysis in Section~\ref{subsec:overall-ranking}, we now explore the sensitivity of the proposed CSD framework to \emph{score perturbation}.

\textbf{Sensitivity to extremely low scores.}\quad 
In real-world scenarios, model performance scores may be subject to significant measurement errors or API instability, leading to occasional extreme (outlier) values on one or several specific benchmarks. 
This analysis investigates the robustness of the CSD framework when faced with such score perturbations.

\begin{figure}[t]
    \centering
    \begin{subfigure}[b]{0.49\textwidth} 
        \centering
        \includegraphics[width=\linewidth]{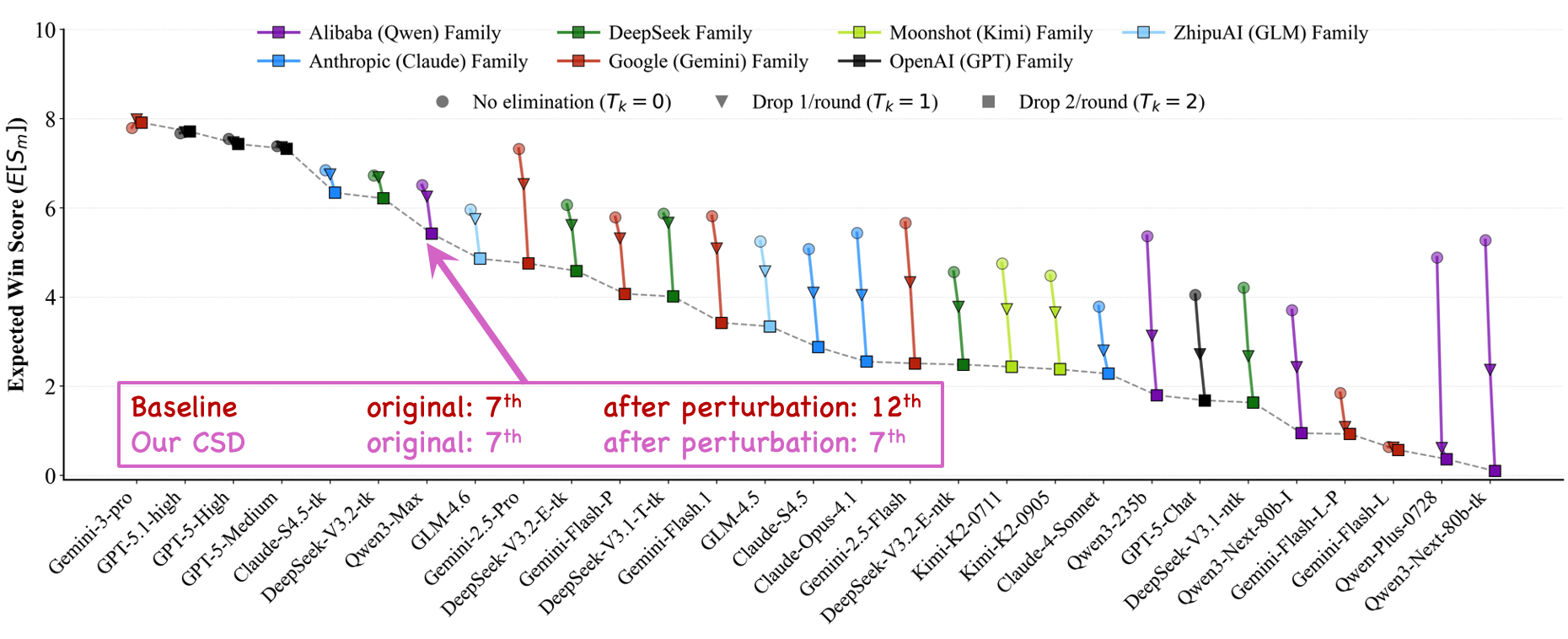}
        \caption{Example 1. Zero scores on two benchmarks.}
        \label{fig:sensitivity-a}
    \end{subfigure}
    \hfill
    \begin{subfigure}[b]{0.49\textwidth}
        \centering
        \includegraphics[width=\linewidth]{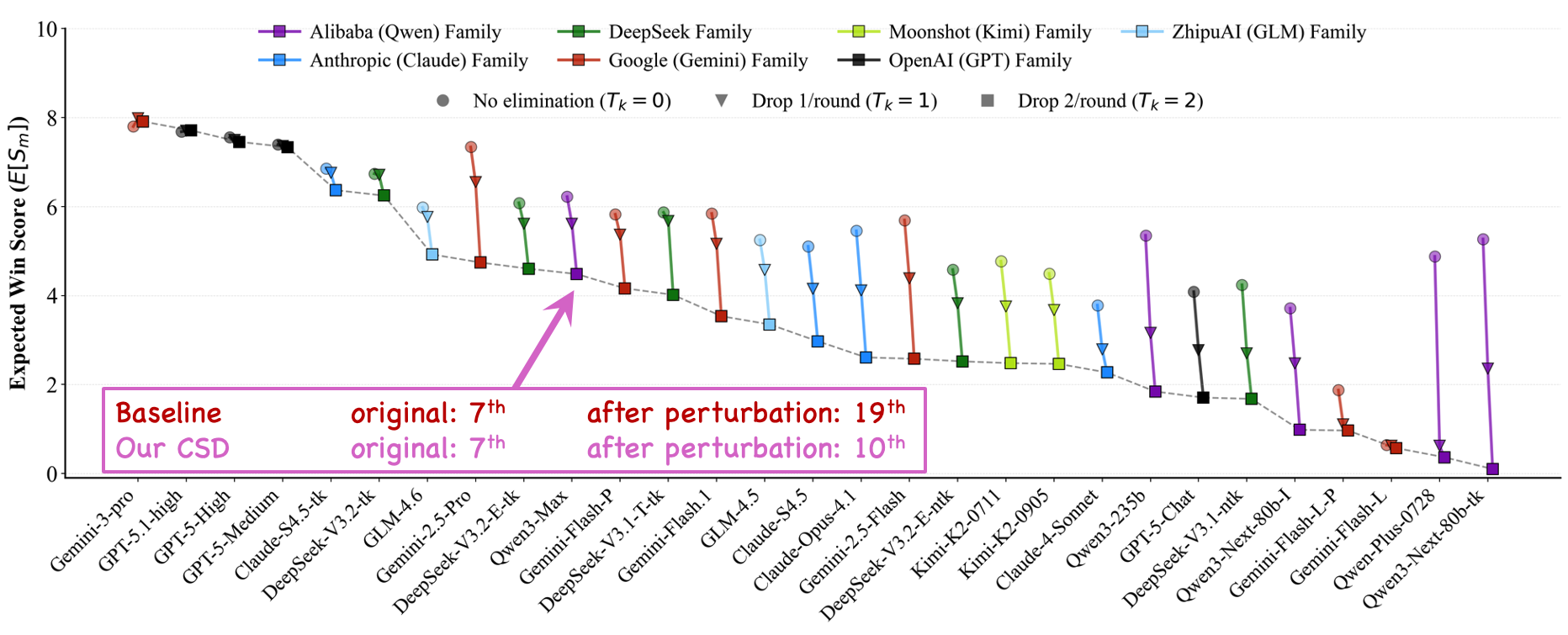}
        \caption{Example 2. Zero scores on four benchmarks.}
        \label{fig:sensitivity-b}
    \end{subfigure}
    \caption{Score Perturbatioin Analysis on Extremely Low Scores.}
    \label{fig:sensitivity-combined}
\end{figure}

\begin{example}[Sensitivity to Zero Scores on IFEval and MulDimIF]
\label{example1}
    Consider a scenario where the scores for Qwen3-Max on the IFEval~\cite{zhou2023instruction} and MulDimIF~\cite{ye2025multi} benchmarks are set to zero, potentially simulating API errors. In this case, Qwen3-Max becomes the weakest model in the third round.
    
    Surprisingly, as depicted in Figure~\ref{fig:sensitivity-a}, this \emph{extreme perturbation minimally affects the overall ranking} produced by our CSD framework.
    In sharp contrast, if we were to simply aggregate all benchmark results using the average score, the rank of Qwen3-Max would drop to $12^{\text{th}}$, highlighting the CSD framework's superior robustness to outlier data.
\end{example}

\begin{example}[Sensitivity to Zero Scores on Four Benchmarks]
\label{example2}
    To further test resilience, we introduce an extreme perturbation by setting the Qwen3-Max scores to zero on four distinct benchmarks: IFEval~\cite{zhou2023instruction}, MulDimIF~\cite{ye2025multi}, AIME24, and AIME25 (simulating a comprehensive failure due to API errors). 
    This places Qwen3-Max as the weakest model in the third round and a significantly weaker model in the sixth.
    
    As detailed in Figure~\ref{fig:sensitivity-b}, under this severe perturbation, the rank of Qwen3-Max drops to $10^{\text{th}}$ but remains within the third performance tier. Crucially, the simple average baseline ranking for Qwen3-Max plummets much further, to $19^{\text{th}}$, underscoring the CSD framework's superior stability against widespread score anomalies.
\end{example}

\textbf{Remark on CSD Robustness}.\quad 
Comparing Example~\ref{example1} with Example~\ref{example2}, we observe that increasing the number of zeroed benchmark scores leads to a more significant, yet controlled, decrease in the model's rank. 
This analysis demonstrates that the CSD framework is robust against isolated, severe score perturbations (e.g., API errors affecting a few benchmarks).
Crucially, the purpose of this sensitivity test is not to validate the framework's behavior under widespread data failure. 
In a practical scenario where a large number of benchmark scores are unreliable, the only necessary step is to re-test all results.
Furthermore, we find that the CSD framework similarly exhibits low sensitivity to extremely high scores on several specific benchmarks.

\section{Discussion}
In this section, we discuss the extended applications of our proposed CSD framework, as well as the limitations of the framework.

\subsection{Extended Applications}
Beyond generating an overall ranking, the CSD framework is flexible and can support personalized and specialized use cases. 
This section discusses two such applications: agentic performance prediction and ranking models within a single benchmark.

\subsubsection{Agentic Performance Prediction}
Our CSD framework is ideally suited for \emph{agentic performance prediction} because its core mechanism requires a \emph{sequence of input benchmarks}. 
This structure directly aligns with the widely-used agentic workflow, where a single complex task necessitates the sequential and competitive execution of multiple underlying capabilities.
For example:

\begin{example}[Web Navigation and Data Extraction]
    A complex agentic task, such as Web navigation and data extraction, involves a sequence of dependent steps where failure in an early step invalidates the entire process. This can be mapped to a competitive sequence for CSD as follows:
    \begin{enumerate}
        \item B1: IFEval (Instruction Following): \textit{Interpret the user's goal and constraints.} (Foundational)
        \item B2: GSM8K (Simple Reasoning): \textit{Devise the initial action plan (e.g., Search $\to$ Click $\to$ Extract).} (Core Planning)
        \item B3: ToolBench (Function Calling): \textit{Execute the action by correctly generating the tool/API interaction code.} (Execution)
        \item B4: HumanEval (Code Debugging): \textit{Validate and process the retrieved data for final output.} (Refinement)
    \end{enumerate}
    Therefore, if we set the corresponding sequence of benchmarks, the overall ranking given by our CSD framework could reflect (or ``predict'') the ranking on the new agentic task.
\end{example}

We believe that agentic performance prediction represents an interesting and important extension direction for our CSD framework.

However, a significant caveat exists: since many recent advanced LLMs inevitably \emph{target or optimize} for performance on open-source agentic benchmarks during their development, making it more challenging to directly apply the CSD framework to ``predict'' performance on novel agentic tasks. 
This occurs because the input benchmarks themselves may be subject to data contamination or overfitting.

\subsubsection{Ranking Models within One Single Benchmark}
While traditional evaluation relies on a single aggregate score (e.g., accuracy), this often fails to capture the competitive dynamics across the different capabilities required within a single, complex benchmark. Our CSD framework can provide a more \emph{nuanced and robust ranking} by modeling the internal structure of the benchmark.

\emph{Establishing Difficulty Tiers}.\quad 
To apply CSD, we first leverage the empirical performance of all models to objectively partition the benchmark's question set into sequential difficulty tiers. Specifically, we can rank all questions by their average model performance (success rate) and group them into ordered subsets: $B_1$ (Easiest), $B_2$ (Medium), $B_3$ (Hardest), and so forth.

\emph{Simulating Competitive Sequence.}\quad
We then treat this sequence of difficulty tiers ($B_1, B_2, \ldots, B_n$) as the input to the CSD framework. This simulates a sequential evaluation where models must successfully ``pass'' the easier challenges before proceeding to the harder ones.

\emph{Favoring Robustness over Spikes.}\quad 
This method inherently favors \emph{robust generalists} over models that exhibit large performance variance. Our expectation is that a model failing on empirically easy questions (low score on $B_1$) but succeeding on complex ones (high score on $B_3$) suggests some instability or unreliable reasoning, possibly due to stochasticity or data contamination. By applying the CSD mechanism, such models will incur a severe performance drop ($\Delta E[S_m]$), resulting in a lower final rank. Conversely, models that maintain consistent performance across the increasing difficulty tiers are rewarded for their stability and resilience.

\textbf{Experiment Setting.}\quad 
We conduct experiments on QA datasets (MMLU-pro~\cite{wang2024mmlu} and SuperGPQA~\cite{team2025supergpqa}), respectively.
For each benchmark, we conduct the following workflow:
    \begin{enumerate}
        \item \textbf{Data-Driven Segmentation:} Instead of using the datasets separately, we group all questions into ten tiers ($B_1$: Questions with $>90\%$ average accuracy, $B_2$: $70\%-80\%$, \dots, $B_{9}$: $<10\%$).
        \item \textbf{CSD Input:} We use $B_1$, $B_2$, \dots, $B_{10}$ as the sequential input.
    \end{enumerate}

\begin{figure}[htbp]
    \centering
    \begin{subfigure}[b]{0.49\textwidth} 
        \centering
        \includegraphics[width=\linewidth]{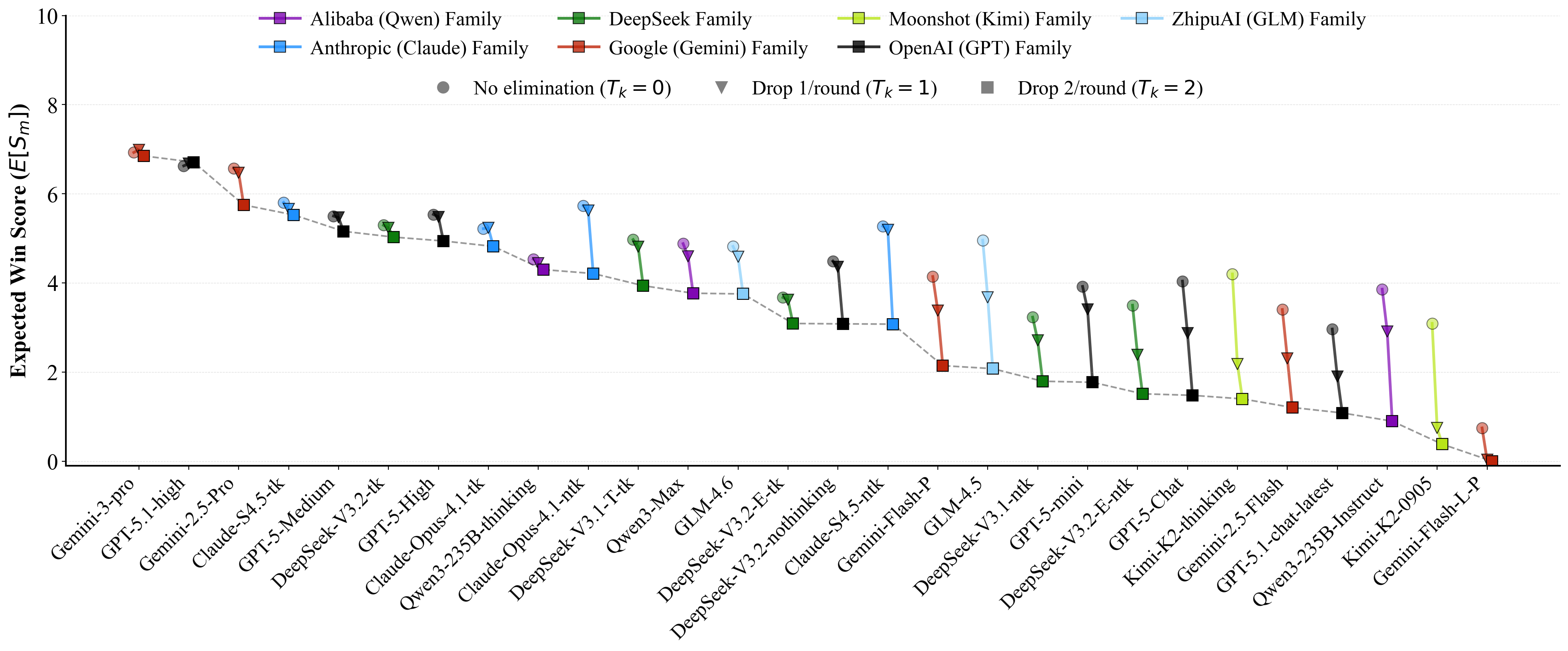}
        \caption{SuperGPQA}
        \label{fig:sensitivity-sgpqa}
    \end{subfigure}
    \hfill
    \begin{subfigure}[b]{0.49\textwidth}
        \centering
        \includegraphics[width=\linewidth]{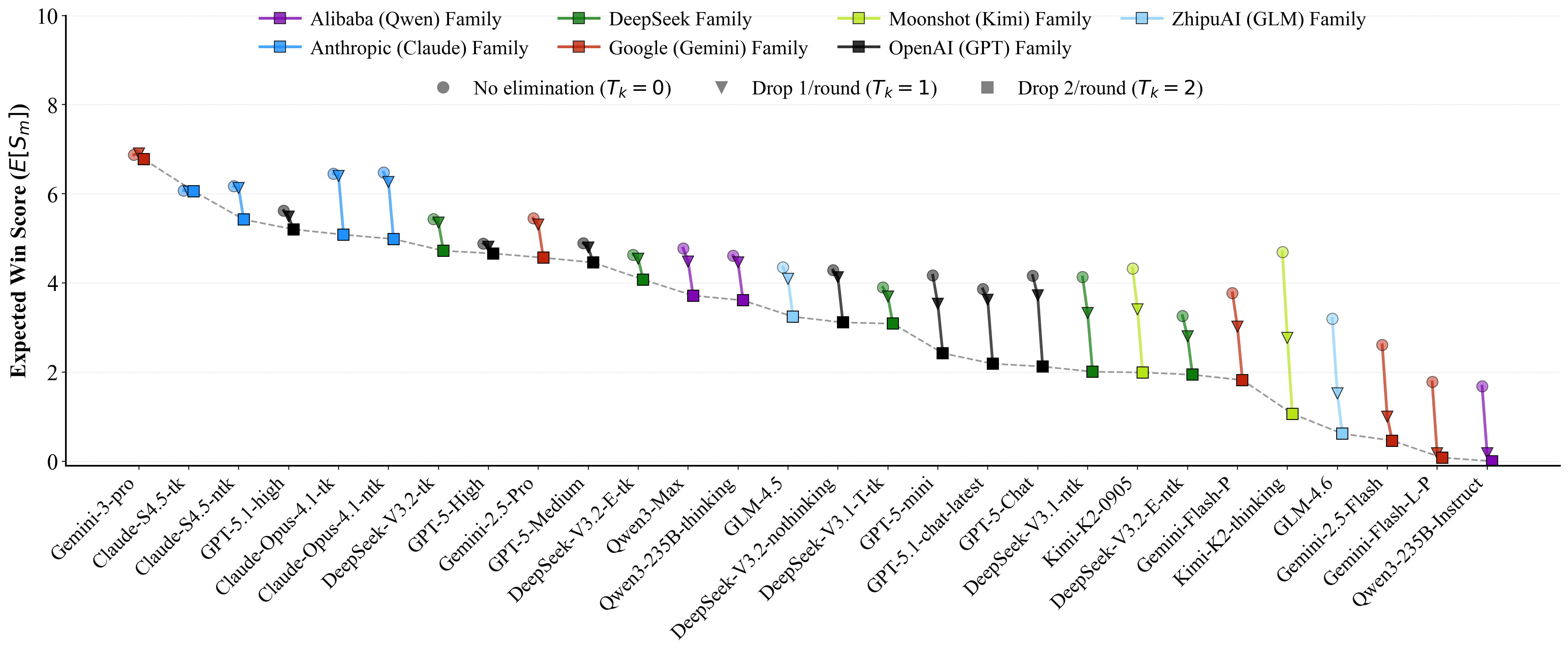}
        \caption{MMLU-pro}
        \label{fig:sensitivity-mmlu-pro}
    \end{subfigure}
    \caption{CSD Framework on Single Benchmark.}
    \label{fig:sensitivity-single}
\end{figure}

As illustrated in~\Cref{fig:sensitivity-single}, models exhibit varying levels of sensitivity when $T_k$ ranges from 0 to 2. Notably, the Kimi models (light green) experience a substantial performance decline on both benchmarks. This suggests that they are less competitive on easier questions compared to harder ones, indicating a lack of robustness across these two benchmarks.
Similar cases happen on Claude nonthinking models and GLM models too.
In comparison, Gemini-3-pro, GPT-5.1-high, and Claude-Sonnet-4.5-thinking are quite robust on these two QA benchmarks.

\subsubsection{Applying CSD to Datasets with Weights}
In cases where only benchmark weights are available without a fixed order, we extend the CSD framework by inducing an order through sampling. Specifically, we sample a permutation of benchmarks based on their weights, ensuring that benchmarks with higher importance are statistically more likely to be listed first.
And based on this sampling mechanism, we can again use Monte Carlo to approximate the expectation of each model's score.
A sample code for generating the benchmark order is like this:

\begin{lstlisting}[language=Python]
    import random
    
    def get_weighted_order(datasets, weights):
        """
        Returns a partial order of datasets based on weights
        using the Efraimidis-Spirakis algorithm.
        """
        # Calculate key: u^(1/w)
        keys = [(random.random() ** (1.0 / w), d) for d, w in zip(datasets, weights)]
        
        # Sort by key in descending order
        keys.sort(key=lambda x: x[0], reverse=True)
        
        return [d for k, d in keys]
\end{lstlisting}

\subsection{Limitations}
While the CSD framework offers significant advantages in identifying robust models and revealing competitive weaknesses, it is essential to acknowledge its limitations, primarily stemming from the nature of LLM evaluation itself.

\textbf{Absence of Ground-Truth Ranking.}\quad 
A fundamental challenge in evaluating complex models is the \emph{absence of a universally accepted ground-truth ranking} for overall LLM performance. Unlike traditional machine learning tasks with clear objective functions, the ``true'' ranking of generalist LLMs remains subjective and context-dependent.
\begin{enumerate}
    \item \textbf{Subjectivity of Utility:} Our CSD ranking reflects the concept of \emph{competitive robustness} and \emph{shortfall penalty}, which favors models that are consistently reliable across diverse tasks. While this is a highly valuable metric for practical deployment, it does not necessarily align with other utility definitions (e.g., peak performance on a single, highly specialized task).
    \item \textbf{Lack of External Validation:} Consequently, it is challenging to perform a definitive external validation of the CSD ranking against an objective ``best'' list. The ranking is primarily validated through its \emph{internal consistency} and its \emph{superior robustness} demonstrated in the sensitivity analysis, rather than by external correlation with an undisputed standard.
\end{enumerate}

\textbf{Challenges in Direct Baseline Comparison.}\quad 
The CSD framework introduces a novel competitive ranking dynamic that diverges significantly from conventional score aggregation methods. This originality presents difficulties when comparing our results to existing baselines.
\begin{enumerate}
    \item \textbf{Incommensurate Metrics:} Traditional baselines often rely on simple metrics like average score or geometric mean, which treat all benchmarks equally and do not account for sequential dependencies or competitive elimination. Since the CSD ranking is defined by its unique $\Delta E[S_m]$ metric (performance drop under competitive pressure), a direct, quantitative comparison with baseline rankings based solely on aggregate scores is incommensurate.
    \item \textbf{Focus on Resilience vs. Peak Performance:} The primary goal of CSD is to penalize weaknesses and reward resilience. Therefore, our ranking may deviate from baselines that prioritize raw peak performance, even if that performance is fragile or susceptible to significant drops when encountering a shortfall. This deviation reflects a deliberate methodological choice rather than an error, but it complicates straightforward ``better/worse'' comparisons with aggregate methods.
\end{enumerate}
\section{Data and Code}
\paragraph{Data.}
We rely on internal evaluation results for this analysis. Please note that LLM scores, especially on agent benchmarks, are sensitive to environmental factors and API stability. 
As a result, the specific rankings derived via the CSD framework may exhibit fluctuations under different evaluation setups.

\paragraph{Code.}
The code can be found at \url{https://github.com/LJSthu/LLMSwissRound}.

\section{Conclusion}
\label{sec:conclusion}

In this paper, we introduced the Competitive Swiss-System Dynamics (CSD) framework, a novel and robust methodology that addresses fundamental limitations in current LLM evaluation paradigms by simulating a dynamic, multi-round competitive environment. This approach intrinsically solves the subjective weighting problem plaguing aggregated leaderboards, as the ``weight'' of a benchmark is naturally determined by its sequence and frequency within the contest structure. By prioritizing dynamic competitive fitness and penalizing performance shortfalls, the CSD ranking offers a more reliable metric for selecting LLMs destined for complex, multi-stage deployment. Future work will focus on extending the CSD framework's utility, including the formal integration of sequential dependencies for agentic performance prediction and exploring the correlation between CSD rankings and real-world task failure rates.

\clearpage

\bibliographystyle{plainnat}
\bibliography{main}

\end{document}